\documentclass[10pt, conference, letterpaper]{ IEEEtran}
\IEEEoverridecommandlockouts
\usepackage{cite}
\usepackage{amsmath,amssymb,amsfonts}
\usepackage{algorithmic}
\usepackage{graphicx}
\usepackage{textcomp}
\usepackage{xcolor}
\usepackage{threeparttable}

\usepackage{amsmath}
\usepackage{amsthm}
\usepackage{multirow}
\usepackage{subfigure}
\usepackage{enumitem}
\usepackage[ruled,vlined]{algorithm2e}  
\usepackage{ragged2e} 
\usepackage{booktabs,makecell, multirow, tabularx}

\usepackage{bm}  

\def\BibTeX{{\rm B\kern-.05em{\sc i\kern-.025em b}\kern-.08em
    T\kern-.1667em\lower.7ex\hbox{E}\kern-.125emX}}

\begin{document}

\title{\textit{Sentinel}: Scheduling Live Streams with \\  Proactive Anomaly Detection 
in \\ Crowdsourced Cloud-Edge Platforms
}
\author{
    Yuting Li\IEEEauthorrefmark{2}, Shaoyuan Huang\IEEEauthorrefmark{2}, Tengwen Zhang\IEEEauthorrefmark{2}, Cheng Zhang\IEEEauthorrefmark{3},  Xiaofei Wang\IEEEauthorrefmark{2}\IEEEauthorrefmark{1}, Victor C.M. Leung\IEEEauthorrefmark{4}\\
    \IEEEauthorblockA{\IEEEauthorrefmark{2}College of Intelligence and Computing, Tianjin University, Tianjin, China} 
    \IEEEauthorblockA{\IEEEauthorrefmark{3}Faculty of Digital Economics and Managements, Tianjin University of Finance and Economics, Tianjin, China} 
    \IEEEauthorblockA{\IEEEauthorrefmark{4}College of Computer Science and Software Engineering, Shenzhen University, Shenzhen, China} 
    Email: \{2020214010, hsy\_23, 2023244126\}@tju.edu.cn,
     zhangcheng@tjufe.edu.cn, xiaofeiwang@tju.edu.cn, vleung@ieee.org
     \thanks{\IEEEauthorrefmark{1}{Corresponding author: Xiaofei Wang.}}
}

\maketitle

\begin{abstract}
With the rapid growth of live streaming services, Crowdsourced Cloud-edge service Platforms (CCPs) are playing an increasingly important role in meeting the increasing demand. 
Although stream scheduling plays a critical role in optimizing CCPs' revenue, 
most optimization strategies struggle to achieving practical results due to the various anomalies in unstable CCPs.
Additionally, the substantial scale of CCPs magnifies the difficulties
of anomaly detection in time-sensitive  scheduling.
To tackle these challenges, this paper proposes Sentinel, a proactive anomaly detection-based scheduling framework. 
Sentinel models the scheduling process as a two-stage Pre-Post-Scheduling paradigm: in the pre-scheduling stage, Sentinel conducts anomaly detection and constructs a strategy pool; 
in the post-scheduling stage, upon request arrival, it triggers an appropriate scheduling based on pre-strategy to implement the scheduling process. 
Extensive experiments on realistic datasets show that Sentinel significantly reduces the anomaly frequency by 70\%, improves revenue by 74\%, and 2.0$\times$ the scheduling speed.



\end{abstract}


\section{INTRODUCTION} 

With the proliferation of mobile devices and advancements in video generation software and hardware, live streaming has gained significant traction in recent years.
The advent of Web3 has not only transformed content production and consumption modes but also facilitated the integration of PGC (Professionally Generated Content), UGC (User-Generated Content), and AIGC (AI-Generated Content) \cite{AIGC}, leading to unprecedented prosperity in the live streaming ecosystem. 

\begin{figure}[tbp]
    \centering
    \includegraphics[width=\columnwidth]{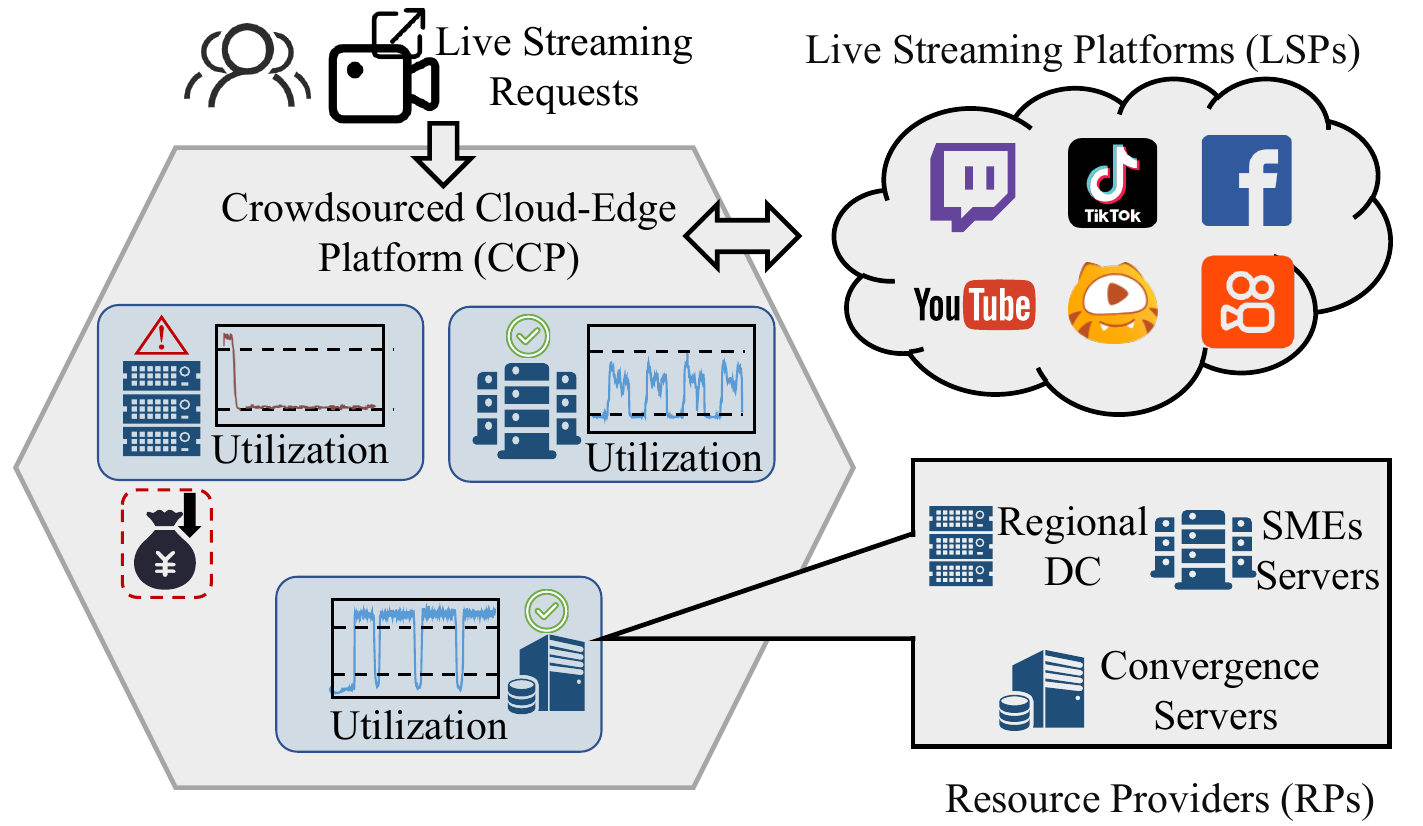}
    \vspace{-1.8em}
    \caption{Crowdsourced Cloud-edge service Platform.}
    \vspace{-1.5em}
    \label{fig:CCP}
\end{figure}

The increasing demand for live streaming has led to the emergence of Crowdsourced Cloud-edge service Platforms (CCPs). As shown in Fig. \ref{fig:CCP}, CCPs support live streaming platforms (LSPs) by utilizing idle resources (e.g., bandwidth) from resource providers (RPs) \cite{xu2021cloud}. Notably, efficient scheduling of live streaming requests is crucial for CCPs' Platform-as-a-Service (PaaS) offering and overall optimization \cite{9355041}.

Despite its importance, scheduling within CCPs is not that easy to be effective. As a matchmaking commercial platform that consolidate a substantial amount of resources, CCPs strive to increase revenue by fully utilizing resources \cite{huang2024seer}. 
We conduct a large-scale measurement study on real-world
traces from a leading CCP.
Fig. \ref{fig:cdf}(a) shows the cumulative distribution function (CDF) of the real-time and daily average utilization of all servers, indicating that a majority of servers experience remarkably low utilization (with 75\% servers' utilization are lower than 20\%).
Fig. \ref{fig:cdf}(b) reveals that there are two prominent peaks within a day and wastage at idle times.
The findings highlight the potential for optimizing the utilization of CCP’s servers through scheduling.

Several recent studies \cite{nasiri2023scheduling, rajab2021iot, zhao2024cur, huang2024seer} have attempted to optimize throughput and revenue through advanced strategies or algorithms, 
but these efforts are limited to the ideal assumption, i.e., CCPs have the ability to provide stable service all the time. 
This is impractical, as sudden drops to zero in Fig. \ref{fig:cdf}(b) suggest potential server failures.
The major live streaming failure during the World Cup in Australia \cite{world2018} underscores that the surge in live streaming services has also led to increased challenges in service capability, which hampers revenue growth and can be attributed to the inherent third-party nature of CCPs: 

($i$) \textbf{Instability of Services.} On one hand, there are a large number of heterogeneous servers of varying quality in CCPs, most are built on hardware not controlled by the platform \cite{huang2023one}. 
These servers fail more frequently than datacenter-level machines \cite{feng2023large}. On the other hand, the distributed deployment of CCPs' servers complicates their operation and maintenance. 

($ii$) \textbf{Scheduling Invalid.} While previous scheduling optimization efforts have been considered effective, they have all overlooked anomalous disruptions, which can only get unrealistic results because the platform will not remain perfectly healthy all the time as people ideally.

Therefore, it is crucial to identify various anomalies within the unstable CCPs,  develop specialized technology to detect them, and integrate these detection into the actual scheduling process. 
However, direct implementation faces two challenges:

\begin{figure}[t]
    \centering
    \includegraphics[width=\columnwidth]{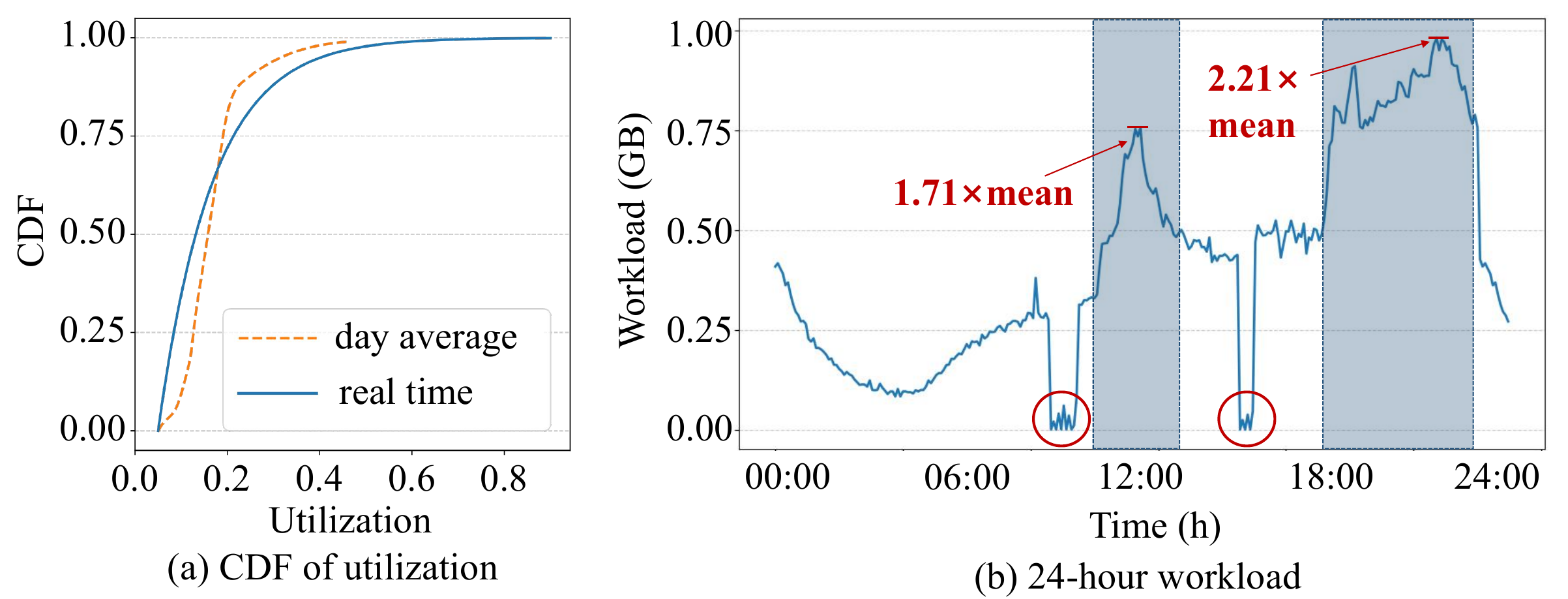}
    \vspace{-1.8em}  
    \caption{Server utilization analysis.}
    \vspace{-1.5em}  
    \label{fig:cdf}
\end{figure}

$(i)$ CCPs manage numerous heterogeneous servers and diverse live streaming requests simultaneously within a complex network environment. This results in multi-dimensional anomalies, each with distinct causes and effects.
$(ii)$ The time consumed in generating and executing scheduling policies is critical, which is often overlooked in prevailing research. The time cost in adding anomaly detection to the process is formidable in a time-sensitive real scheduling process.


To tackle these challenges, we propose a comprehensive solution that focuses on refining the scheduling process to detect anomalies and augment CCPs revenue.
By analyzing the anomalies of the real-world traces from a leading cloud-edge provider, we identify the major anomalies that most impair the platform. 
Moreover, 
we model the relationship between resource utilization and revenue under the influence of
diverse anomalies in a unified metric.

Ultimately, we introduce Sentinel, a proactive anomaly detection-based live streaming scheduling system for revenue optimizing.
Rather than relying on reactive troubleshooting, 
Sentinel adopt proactive approach to avoid
invalid scheduling.
Specially, through extending the native scheduling framework into a two-stage \textbf{Pre-Post-Scheduling} (\(\mathrm{P^2S}\)) paradigm, 
Sentinel enables customized anomaly detection prior to the time-sensitive, real-time scheduling process, reducing decision latency and attaining revenue optimization. Our contributions are summarized as follows:





\begin{itemize}[leftmargin=*]
\item To the best of our knowledge, we are the first to focus on the revenue optimization of CCPs serving the emerging live streaming industry in the face of multi-dimensional anomalous disruptions.
\item Utilizing a comprehensive dataset from real-world CCP operations, we observe and analyze a vast number of scheduling incidents, which allows us to identify and detect critical features that lead to anomalies. 
We also devise a uniform metric that delineates the relationship between resource utilization and revenue.
\item We introduce Sentinel, a proactive anomaly detection-based scheduling system, which combines anomaly detection and scheduling optimization through a novel \(\mathrm{P^2S}\) paradigm.  
Evaluations based on real-world data demonstrate that  compared to its counterparts, Sentinel significantly reduces the anomaly frequency by 70\% , improves revenue by 74\%, and $2.0\times$ the scheduling speed.
\end{itemize}

\begin{figure*}[htbp]
    \centering
    \includegraphics[width=\textwidth]{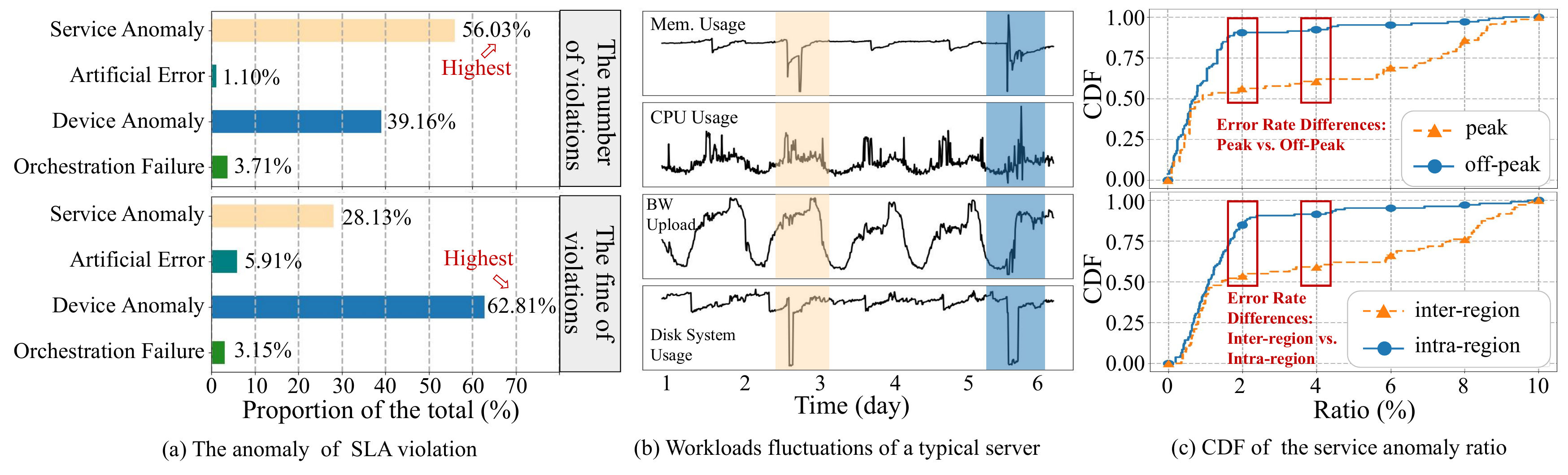}
    \vspace{-2.0em}  
    \caption{Analysis of the anomalies.}
    \vspace{-1.5em}  
    \label{fig:Anomaly Analysis}
\end{figure*}

\section{MOTIVATION}\label{npd}
Our research is
conducted in collaboration with a leading CCP in China, which
consists of 5174 servers distributed across the country and
serves dozens of typical LSPs.     
Details of datasets can be found in Sec. \ref{exp_setting}.

\subsection{Identifying Anomalies in CCPs}
The primary motivation for this work is the observation of various anomalies within CCPs that disrupt requests scheduling, with an observed anomaly rate of approximately 16.1\% in the whole CCP. 
Given that multiple anomalies cause Service Level Agreements (SLA) violations, our anomaly analysis focuses on two key metrics: (i) the number of anomalies attributed to each factor, which aims to identify those most likely to cause SLA violations, and (ii) the number of fines resulting from anomalies, which seeks to identify the anomalies posing the greatest threat to revenue.

Due to the confidentiality of SLAs as proprietary business information, we generalize the causes of violations into four categories: \textbf{Service anomaly}, covering network connections and communications; \textbf{Artificial error}, including adjustments and testing by engineers; \textbf{Device anomaly}, related to hardware failures; and \textbf{Orchestration failure}, involving container configuration and operation.

As depicted in Fig. \ref{fig:Anomaly Analysis}(a), service-related issues are the primary cause of SLA violations, accounting for 56.03\% of the overall violations and 28.13\% of fines. However, device-related issues, although only 39.16\% of violations, result in the highest fines, contributing to 62.81\% of the total fines. This can be attributed to the fact that device anomalies render most server resources unavailable, preventing proper request processing, whereas service anomalies typically only disrupt the current request. Based on this analysis, we focus on two types of anomalies: \textbf{device anomaly} and \textbf{service anomaly}.

\subsection{Characterizing Device Anomaly and Service Anomaly}\label{anomaly}
In this part we characterize these two anomalies further to provide guidance for subsequent detection.

\textbf{Device Anomaly.}
While many anomaly detection methods suggest relying on fluctuations within monitoring series \cite{ren2019time}, this approach is insufficient for CCPs due to their non-stationary temporal characteristics caused by user behavior and scheduling. These normal fluctuations can be mistaken for device failures, leading to numerous false positives and low detection accuracy.

Fig. \ref{fig:Anomaly Analysis}(b) illustrates the four resource workloads for a typical server experiencing  device anomaly, While significant fluctuations are observed in both the yellow and blue segments, they represent healthy and anomaly states, respectively.
In the yellow segments, the memory usage and disk usage exhibit normal fluctuations with consistent tendency. Nevertheless, there are sudden changes with inconsistent tendency of each series in the blue segments, which can be attributed to correlation disturbances caused by device anomaly. Consequently, it is vital to consider both non-stationary temporal characteristics (fluctuations) and correlation disruptions (inconsistent patterns) when detecting device anomaly. 


\textbf{Service Anomaly.}
We identify two key factors that significantly impact service anomalies from temporal and spatial perspectives: request timing (peak vs. off-peak) and request location (inter-region vs. intra-region).

As shown in Fig. \ref{fig:Anomaly Analysis}(c), the CDF of the service anomaly rate under different conditions reveals a substantial separation between the lines in each subplot, indicating that both factors significantly influence the occurrence of service anomalies.
On one hand, service anomaly predominantly occur during peak times, which is probably due to the increase in network load and intense  competition for resources during the peak period \cite{huang2024seer}.
On the other hand, inter-region scheduling is more prone to service anomalies, which can be attributed to the fact that cross-region scheduling involves more complex network paths and potential network delays inherent in cross-region operations \cite{zhang2022aggcast}. Therefore, we choose to leverage these two features to provide guidance for service anomaly detection.
\subsection{Heterogeneous Resource Revenue Metric}
In this section, we aim to establish a unified revenue metric for CCPs heterogeneous servers, especially considering the interference of aforementioned anomalous factors. 

\begin{figure}[t]
    \centering
    \includegraphics[width=\columnwidth]{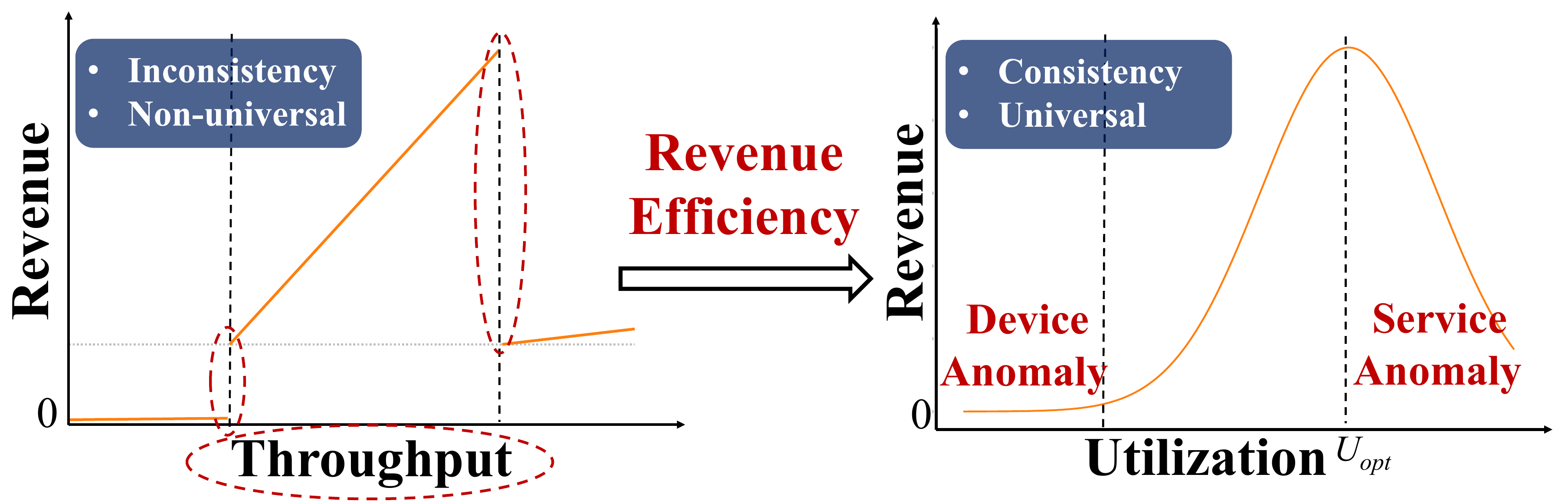}
    \vspace{-1.5em}  
    \caption{Resource revenue metric.}
    \vspace{-1.8em}  
    \label{fig:line}
\end{figure}

Ideally, server revenue can be obtained directly by multiplying throughput $T$\protect\footnotemark[1] by unit price $F_p(\cdot)$ as $R = F_p(T)$.
\footnotetext[1]{Given the nature of live streaming services, all references to ’throughput’ and its derived notions
in the following sections specifically denote bandwidth.} Due to the fines from SLA violations imposed by anomalies, this relationship is not simple linear.
By measuring more than 300 servers in reality, we observe a complex nonlinear relationship, as shown in Fig. \ref{fig:line}. 
This throughput-revenue curve exhibits sudden changes at both low and high throughput levels due to device and service anomalies, respectively.


However, it can hardly be used directly as the CCPs revenue metric due to its \textbf{discontinuities} and \textbf{non-universality} (servers with different hardware conditions exhibit varying throughput levels and prices).
To address these issues, we design a unified heterogeneous resource revenue metric for Sentinel. We employ \textbf{utilization} $\bm{U}$
as a relative metric to evaluate the revenue across heterogeneous servers, and propose \textbf{revenue efficiency} $\bm{F_r(\cdot)}$ to model the relationship as $R = F_r(U)$.

On this basis, we model the consistency of the revenue metric by incorporating the probability of anomaly occurrence as a discount to revenue. Specially, we introduce a Gaussian function to model $F_r(\cdot)$ and implement smoothing as:
\vspace{-0.5em} 
\begin{equation}
R(U)= e^{-\frac{\left(U - U_{o p t}\right)^2}{2 \sigma^2}},
\vspace{-0.5em} 
\end{equation}
where $U$ is the utilization of arbitrary server, $U_{o p t}$ is the optimal utilization (detailed in Eq. \ref{eq: u_opt}), and $\sigma$ reflects the sensitivity of revenue to anomalies.

As shown in Fig. \ref{fig:line}, when utilization is low, revenue is negligible due to potential device anomalies. As utilization increases, revenue rises rapidly due to enhanced processing capabilities and economies of scale. However, when utilization exceeds $U_{o p t}$, network load increases, leading to higher SLA violation fines from service anomalies during peak periods. By introducing revenue efficiency, we establish a unified resource revenue metric for heterogeneous servers.




\begin{figure*}[h]
    \centering
    \includegraphics[width=\textwidth]{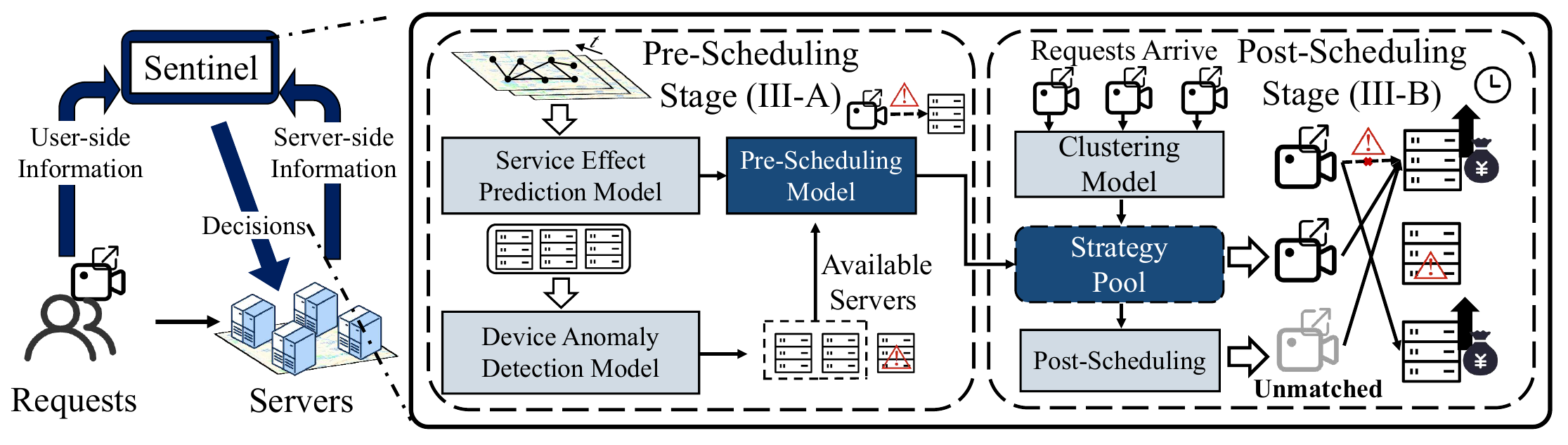}
    \vspace{-1.8em}  
    \caption{The system overview of Sentinel.}
    \vspace{-1.3em}  
    \label{fig:Sentinel}
\end{figure*}

\section{METHODOLOGY}\label{SYSTEM}
Fig. \ref{fig:Sentinel} provides the overview of Sentinel, a proactive anomaly-aware live streaming scheduling system. Sentinel detects device anomalies across the entire platform and service anomalies for each live request, while optimizing scheduling decisions to maximize CCP revenue.
Sentinel functions by extending the native scheduling into a \textbf{Pre-Scheduling} stage and a \textbf{Post-Scheduling} stage with \textbf{Strategy Pool}.


To address the significant time costs of detection in real-time scheduling, Sentinel conducts anomaly detection during the pre-scheduling stage before the actual requests arrive, generating revenue-optimized decisions according to the detection results and feeding them into the strategy pool.

When actual requests arrive, Sentinel matches and schedules them according to the strategy pool. Any untreated requests due to prediction errors are handled in the post-scheduling stage through a simple and efficient strategy.

\subsection{Pre-scheduling Stage with Anomaly Detection}
As the core stage of Sentinel, the pre-scheduling stage is composed of device anomaly detection, service effect prediction, and revenue-optimised scheduling.

\subsubsection{Device Anomaly Detection}
\begin{figure}[tbp]
    \centering
    \includegraphics[width=\columnwidth]{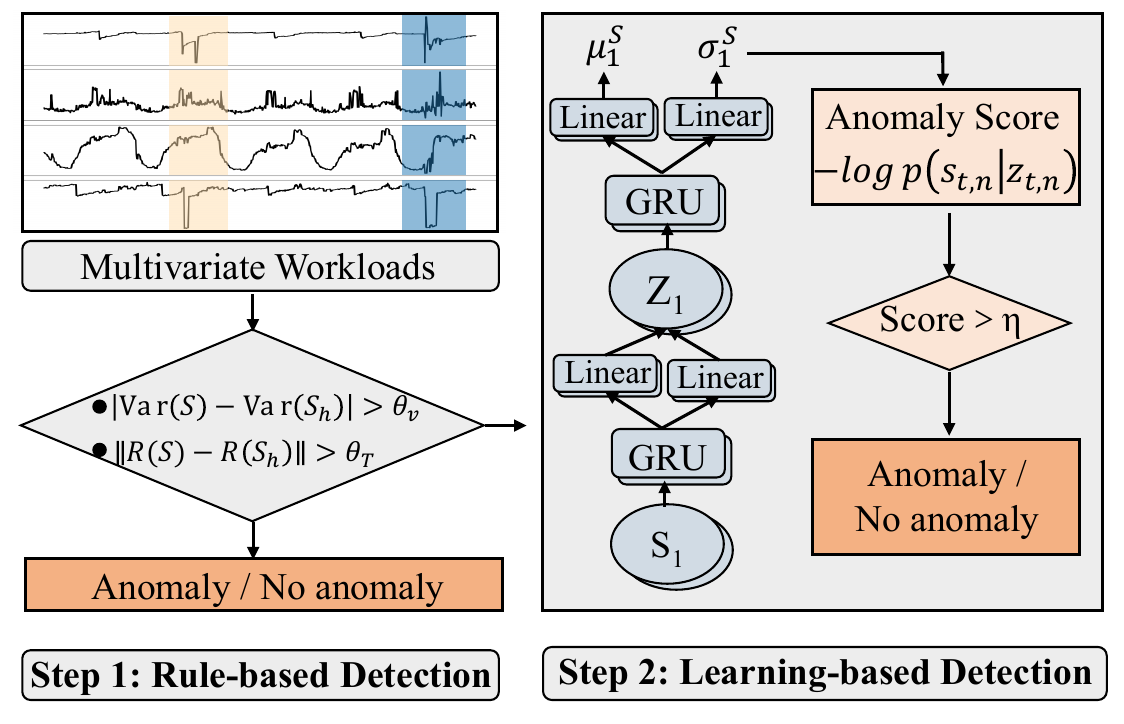}
    \vspace{-1.8em}  
    \caption{Two stage anomaly detector.}
    \vspace{-1.3em}  
    \label{fig:de anomaly}
\end{figure}

In this section, we design a two-step device anomaly detector, consisting of rule-based detection and learning-based detection (Fig. \ref{fig:de anomaly}). 
Both methods are tailored to detect non-stationary temporal characteristics and inconsistent workload patterns.

\textbf{Step 1: Rule-based Detection.} 
Inspired by the analysis in Sec. \ref{anomaly}, the rule-based detection applies manual conditions to decide whether the server is down based on the fluctuations of each hardware workload and the correlations between them. 

Sentinel monitors the server workloads using a slide window $S$ with a length of $T$, with the multivariate workload series with $N$ dimensions as $\mathcal{S} =\{s_{t, n} | t \in [1, T], n \in [1, N]\}$. 
Sentinel also selects a record of series in a healthy condition $\mathcal{S}_h$ as a baseline.
Given the heterogeneity of servers, the baseline is meticulously selected from devices that deliver analogous types of services and exhibit similar operational scales. 
The rule-based detection follow two conditions:
\begin{itemize}[leftmargin=1em]
 \item  Comparing $\mathcal{S}$ and $\mathcal{S}_h$, judging the difference between the variance of $\beta$ dimensions is over $\theta_v$.
\item Computing the correlation matrices between $\mathcal{S}$ and $\mathcal{S}_h$, respectively, judging the difference between the matrix norms exceeds $\theta_r$.
\end{itemize}
The rule-based detection can fast filter some failure servers, which is particularly valuable in second-by-second scheduling.

If both conditions are met, the server is deemed faulty and excluded from subsequent scheduling; if neither condition is met, the server is considered available. When only one condition is met, the rule-based detector struggles to make a clear judgment. In such case, we employ a learning-based detection approach, detailed in the following section.

\textbf{Step 2: Learning-based Detection.}
The learning-based detection applies a deep learning model based on Variational Gated Recurrent Unit (VGRU) \cite{chung2015recurrent}, consisting of two main parts: the inference stage and the generation stage.

At inference stage, the encoder maps the original input $s_{t,n}$ and the history information ${h}_{t-1, n}$ from GRU to the latent random variable $\boldsymbol{z}_{t, n}$ as: 
\vspace{-0.3em} 
\begin{equation}\label{eq:enc}
\begin{aligned}
& {\boldsymbol{z}}_{t, n} \mid \boldsymbol{s}_{t, n} \sim \mathcal{N}\left(\boldsymbol{\mu}_{t, n}^z, \boldsymbol{\sigma}_{t, n}^z\right), \\
& \boldsymbol{\mu}_{t, n}^z=\varphi_\mu^{\operatorname{enc}}\left(\varphi^s\left(\boldsymbol{s}_{t, n}\right), \boldsymbol{h}_{t-1, n}\right), \\
& \boldsymbol{\sigma}_{t, n}^z=\varphi_\sigma^{\operatorname{enc}}\left(\varphi^s\left(\boldsymbol{s}_{t, n}\right), \boldsymbol{h}_{t-1, n}\right),
\end{aligned}
\vspace{-0.3em} 
\end{equation}
where $\boldsymbol{\mu}_{t, n}^z$ and $ \boldsymbol{\sigma}_{t, n}^z$ are the mean and variance parameters of
the variational distribution.
$\varphi^{\operatorname{enc}}\left(\varphi^s\left(\boldsymbol{s}_{t, n}\right), \boldsymbol{h}_{t-1, n}\right) = f_\theta({W}_s \cdot \varphi^s(\boldsymbol{s}_{t, n})+{W}_h \cdot \boldsymbol{h}_{t-1, n} + {b}_h)$, $f_\theta$ is a non-linear activation function, and $\varphi^s(\boldsymbol{s}_{t, n})$ represents the the
convolutional network to extract features. 
This process is the posteriori update for ${\boldsymbol{z}}_{t, n}$.

At the generation stage, the latent variable ${\boldsymbol{z}}_{t, n}$  is reconstructed back to the original input ${\boldsymbol{s}}_{t, n}$  through the decoder:  
\vspace{-0.3em} 
\begin{equation}
\begin{aligned}
& \tilde{\boldsymbol{s}}_{t, n} \mid \boldsymbol{z}_{t, n} \sim \mathcal{N}\left(\boldsymbol{\mu}_{t, n}^s, \boldsymbol{\sigma}_{t, n}^s\right), \\
& \boldsymbol{\mu}_{t, n}^s=\varphi_\mu^{\operatorname{dec}}\left(\boldsymbol{z}_{t, n}, \boldsymbol{h}_{t-1, n}\right), \\
& \boldsymbol{\sigma}_{t, n}^s=\varphi_\sigma^{\operatorname{dec}}\left(\boldsymbol{z}_{t, n}, \boldsymbol{h}_{t-1, n}\right),
\end{aligned}
\vspace{-0.5em} 
\end{equation}
where $\boldsymbol{\mu}_{t, n}^s, \boldsymbol{\sigma}_{t, n}^s$ denote parameters of the generating distribution,  
and $\varphi_\mu^{\mathrm{dec}}$ and $\varphi_\sigma^{\mathrm{dec}}$ can also be any nonlinear flexible functions and we also use neural network like Eq. \ref{eq:enc}. 
Finally, to better generate the structures of data, we further apply
a deconvolutional network as $ {\boldsymbol{s}}_{t, n} = \varphi^{\tilde{s}} (\tilde{\boldsymbol{s}}_{t, n})$.

Since the learning-based model is trained to learn the normal
patterns of multivariate workloads, the more an observation follows normal patterns, the more likely it can be reconstructed with higher confidence. Hence, we apply the reconstruction
error of $\mathcal{S}_{t}$, i.e.,   $-\log p\left(\boldsymbol{s}_{t, n} \mid \boldsymbol{z}_{t, n}\right)$, as the anomaly score to determine whether a server is anomalous \cite{su2019robust}:
\vspace{-0.5em} 
\begin{equation}
\begin{cases}\text { anomaly, } & \text { if } -\log p\left(\boldsymbol{s}_{t, n} \mid \boldsymbol{z}_{t, n}\right)>\eta \\ \neg \text { anomaly, } & \text { otherwise, }\end{cases}
\vspace{-0.5em} 
\end{equation}
where $\eta $ denotes the threshold.
An observation $\mathcal{S}_{t}$ will be classified as anomalous if anomaly score is below $\eta $.


However, traditional VGRU models use a single Gaussian distribution for latent variables $\boldsymbol{z}_{t, n}$, making it challenging to capture diverse patterns in heterogeneous servers and unsuitable for large-scale CCPs. 
To address this, we incorporate a probabilistic mixture into VGRU. We assume the latent variables $\boldsymbol{z}_{t, n}$ are drawn from a Gaussian mixture distribution, with each component's parameters including a specific prior for the current input and a transformation of the latent state from the previous timestep.

We define the associated latent discrete variable $\boldsymbol{c}_{t, n}$ for each $s_{t, n}$ to guide the prior selection of the
current timestep.
The prior of the latent variable is given by:
\vspace{-0.5em} 
\begin{equation}
\boldsymbol{z}_{t, n} \mid \boldsymbol{c}_{t, n} \sim \prod_{k=1}^K \mathcal{N}\left(\boldsymbol{z}_{t, n} \mid \boldsymbol{\mu}_k, \operatorname{diag}\left(\boldsymbol{\sigma}_k\right)\right)^{c_{t, n, k}},
\vspace{-0.5em} 
\end{equation}
where $\{\boldsymbol{\mu}, \boldsymbol{\sigma}\}=\left\{\boldsymbol{\mu}_k, \boldsymbol{\sigma}_k\right\}_{k=1}^K$ represent multiple Gaussian distribution and $K$ is the number of components.
The one-hot vector $\boldsymbol{c}_{t, n}=\left(c_{t, n, 1}, \ldots, c_{t, n, K}\right)^T$ follows a categorical distribution $\boldsymbol{c}_{t, n} \sim \operatorname{Cat}(\pi)$ with parameter $\pi \in \mathbb{R}_{+}^{K \times 1}$.

This way allows us to assign a different Gaussian distributed prior conditioned on $\boldsymbol{c}_{t, n}$, capturing the diverse distribution characteristics of the input.  
By marginalizing $c_{t, n}$, we obtain:
\begin{equation}
\begin{aligned}
\boldsymbol{z}_{t, n} &\sim \sum_{c_{t, n}} p\left(\boldsymbol{c}_{t, n} \mid \boldsymbol{\pi}_{t, n}\right) p\left(\boldsymbol{z}_{t, n} \mid \boldsymbol{c}_{t, n}\right) \\
&= \sum_{k=1}^K \pi_{t, n, k} \mathcal{N}\left(\boldsymbol{\mu}_k, \operatorname{diag}\left(\boldsymbol{\sigma}_k\right)\right).
\end{aligned}
\end{equation}
Obviously, it is mixture Gaussian distribution with higher representation power than a Gaussian distribution, and thus is ideal for characterizing the heterogeneous server workloads.

Specially, we employ variational inference to learn $\boldsymbol{c}_{t, n}$. 
However, direct optimization of $\boldsymbol{c}_{t, n}$ is challenging since the back-propagation algorithm cannot be applied to non-differentiable layers.
Inspired by \cite{blei2006variational}, we approximate samples from the categorical distribution using the Gumbel-softmax distribution
$q\left(\boldsymbol{c}_{t, n}\right)= \text{Gumble-softmax} \left(\overline{\boldsymbol{\pi}}_{t, n}\right)$, $\overline{\boldsymbol{\pi}}_{t, n} \in \mathbb{R}^{K \times 1}$ is the parameter for $q\left(\boldsymbol{c}_{t, n}\right)$ and it draws samples via:
\begin{equation}
\begin{aligned}
c_{t, n, k} & =\frac{\exp \left(\left(\log \bar{\pi}_{t, n, k}+g_{t, n, k}\right) / \lambda\right)}{\sum_{k=1}^K \exp \left(\left(\log \bar{\pi}_{t, n, k}+g_{t, n, k}\right) / \lambda\right)}, \\
\text { for } k & =1, \ldots, K \\
g_{t, n, k} & \sim \operatorname{Gumbel}(0,1)=-\log \left(-\log \left(\epsilon_{t, n, k}\right)\right)
\end{aligned}
\end{equation}
where $\lambda$ is the softmax temperature and $\epsilon_{t, n, k}$ is a standard uniform variable. 
The parameter $\overline{\boldsymbol{\pi}}_{t, n}$ for $q\left(\boldsymbol{c}_{t, n}\right)$ is given by $\overline{\boldsymbol{\pi}}_{t, n} = \operatorname{softmax}(\overline{W}_s \cdot \overline{\boldsymbol{s}}_{t, n}+\overline{W}_h \cdot \boldsymbol{h}_{t-1, n} + \overline{b}_h)$.

After obtaining $\boldsymbol{c}_{t, n}$, it is fused with the prior to guide the inference of $\boldsymbol{z}_{t, n}$.  
By sampling from a Gaussian mixture  distribution,  
it efficiently models the diversity patterns.

The meticulously two-stage detector balances detection accuracy and efficiency, generating the anomaly index $\mathbf{D} \in\{0,1\}^{E}$ for each server.
If a device anomaly is detected on server $e$ then $\mathbf{D}_e$ is set to 0; otherwise, it is set to 1.

\subsubsection{Service Effect Prediction}
The service effect prediction consists of two steps: upcoming requests prediction and their service effects modeling, which involve the request revenue (the average data throughput per minute for a request on a given server), and the potential anomaly status.  

\textbf{Step 1: Request Prediction.}
The main obstacle to request prediction is the inability to simultaneously predict multiple heterogeneous characteristics of requests. 
To address this, we incorporate heterogeneous feature information into request categories via clustering \cite{huang2024seer,huang2022fine}, which allowing Sentinel to predict the volume of each category without needing to forecast individual request features, substantially
simplifying the prediction task.
Specifically, Sentinel classify requests into different categories via KMeans based on the live content of the request and the platform from which the request was sent.

Prior to requests arrival, Sentinel first predicts the request category distribution of all $M$ regions: $\mathcal{R}=\{R_{t, m} | t \in [1, T], m \in [1, M] \}$ for $T$ time steps, where $R_{t, m}=\{r_t^1,...,r_t^N\}$ denotes the number of requests of all $I$ categories at time $t$ in region $m$.

\begin{figure}[t]
    \centering
    \includegraphics[width=\columnwidth]{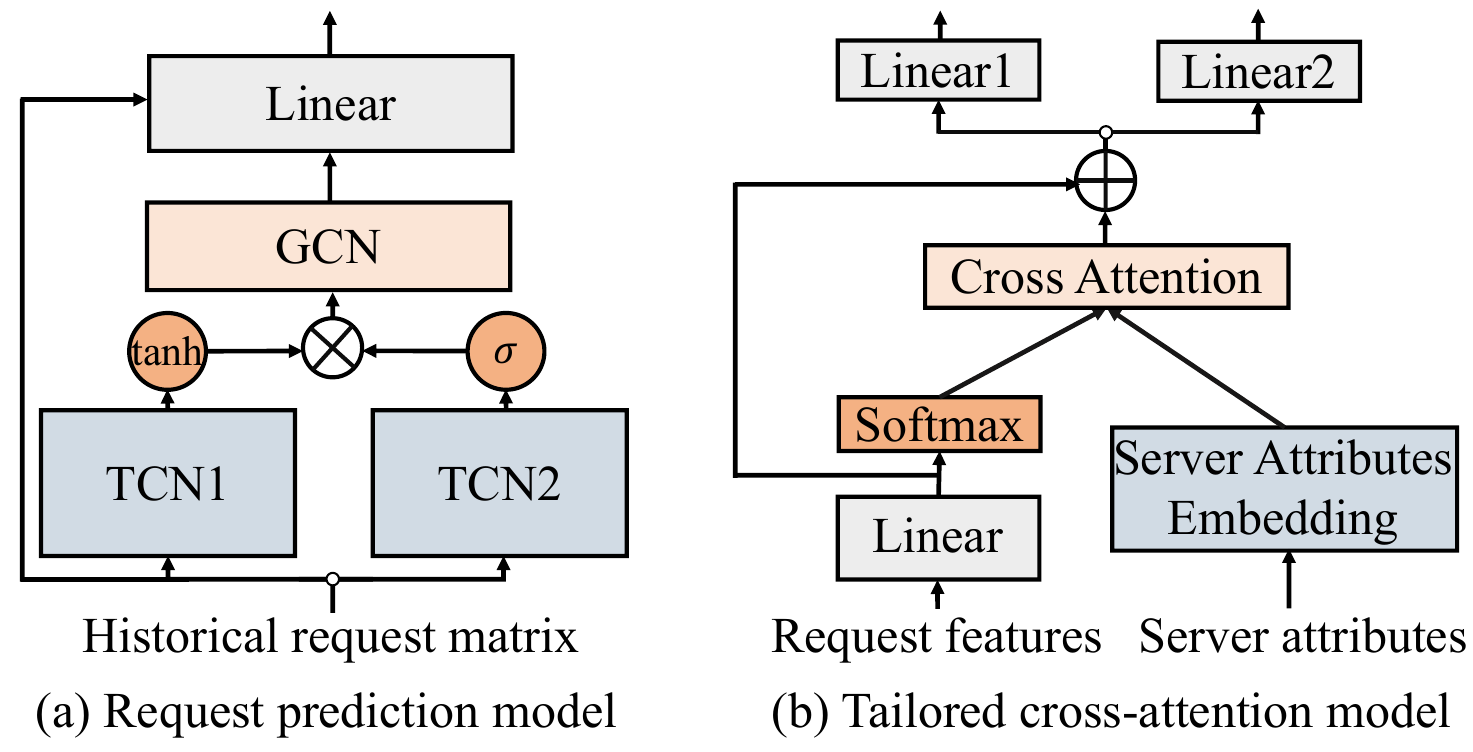}
    \vspace{-1.8em}  
    \caption{Service effect prediction.}
    \vspace{-1.3em}  
    \label{se anomaly}
\end{figure}

We introduce a deep learning model that integrates a gated temporal convolution ($\mathrm {TCN}$) module that composed of two parallel layers with a graph convolutional network ($\mathrm{GCN}$) layer for spatial-temporal feature extraction (Fig. \ref{se anomaly}(a)).   

Specially, we define the user request of all $M$ locations as a graph: $\mathcal{G}=\left(\mathcal{V}, \mathcal{E}, \mathcal{A},\mathcal{R} \right)$ , where $\mathcal{V}$ is the set of spatial locations (nodes) with the size of $|\mathcal{V}|=M$, and $\mathcal{E}$ is a set of edges connecting two spatially adjacent regions in $\mathcal{V}$. The adjacent matrix is denoted as $\mathcal{A} \in \mathbb{R}^{M \times M}$ and is computed in a self-adaptive way to unveil the implicit correlations \cite{wu2019graph}.
We adopt a truncated expansion of $\mathrm{GCN}$ by Chebyshev polynomials to the first order and details are shown as follows:
\vspace{-0.5em} 
\begin{equation}
\begin{aligned}
&\mathcal{H}' ={\sigma}_1\left(\mathbf {TCN_1( \mathcal{R})}\right) \odot {\sigma}_2\left(\mathbf {TCN_2(\mathcal{R})}\right),\\&
\mathcal{H}=\tilde{D}^{-\frac{1}{2}} \tilde{A} \tilde{D}^{-\frac{1}{2}} \mathcal{H}' \Theta,\\&
\hat{\mathcal{R}}_{t+1}={\sigma}_3\left(W_d \cdot \mathcal{H}+b_d\right), 
\end{aligned}
\vspace{-0.5em} 
\end{equation}
where $\odot$ is the element-wise product, $\sigma(\cdot)$ represents the
activation function, we use the tanh, 
sigmoid, and ReLU function here,respectively.
$D$ is the graph degree matrix, we normalize them through the equation $\tilde{A}=A+I_N, \tilde{D}_{i i}=\sum_j \tilde{A}_{i j}$, and $I_N$ is the identity matrix.
Given our use of the standard $\mathrm{TCN}$, for brevity, we adopt $\mathbf {TCN}$ to represent the standard computational process inherent in the $\mathrm{TCN}$.

\textbf{Step 2: Service Effect Modeling.}
After obtaining the $\hat{\mathcal{R}}_{t+1}$, 
Sentinel analyzes their service effect by simultaneously considering two crucial perspectives.
As discussed in Section \ref{anomaly}, the likelihood of service anomalies is influenced by both the intrinsic features of the request and the server attributes, such as inter-region scheduling. Additionally, request revenue is affected by these dimensions; for instance, different bandwidth capacities can lead to varying revenues for the same request.



To effectively integrate and leverage these dual perspectives, we introduce a tailored cross-attention method (Fig. \ref{se anomaly}(b)). This method is designed to dynamically combine the intrinsic features of request with server attributes, ensuring that the server attributes most relevant to the current request's service effect are prioritized.
Specifically, our model uses inputs such as request category, location, timing, server ID, bandwidth, and geographical location, with corresponding request revenue and service anomaly status as labels for training.



To boost prediction efficiency, the service effect model directly generate the the request revenue matrix $\mathbf{A}\in \mathbb{R}^{E \times M \times N}$ and service anomaly matrix $\mathbf{S}\in  \{0,1\}^{E \times M \times N}$ through two specialized decoders:
\vspace{-0.4em}
\begin{equation}
\begin{aligned}
Q,V &= \varphi^{\operatorname{enc_1}}(i,m,t), \varphi^{\operatorname{enc_2}}(e,B_e,L_e),\\
\alpha & =\operatorname{softmax}\left(Q\right), \\
\mathbf{A}_{e,m,i} &= \varphi^{\operatorname{dec_1}}(Q+\operatorname{Dropout}\left(\alpha \cdot V\right)), \\
\mathbf{S}_{e,m,i} &=\varphi^{\operatorname{dec_2}}(Q+\operatorname{Dropout}\left(\alpha \cdot V\right)),\\
&(e\in[1,E], m\in[1,M], i\in[1,I]) 
\end{aligned}
\vspace{-0.4em}
\end{equation}
where $\mathbf{A}_{e,m,i}$ represents the revenue from deploying class $i$ requests from region $m$ to server $e$, $\mathbf{S}_{e,m,i}$ reflects the 
 corresponding service anomaly.
$\varphi^{\operatorname{enc}}$ and $\varphi^{\operatorname{dec}}$ are the nonlinear flexible functions like Eq. \ref{eq:enc}.
$i, m, t, e, B_e, L_e$ are corresponding input features, $E$ represents the number of servers, and $M$ and $I$ are defined as before. 
The cross-attention is not only efficient but also generalized, aiding the of request-server combinations unseen in the historical data and enhancing the robustness of predictions.



\subsubsection{Pre-scheduling Strategy}\label{prescheduling}
With the predicted request matrix $\hat{\mathcal{R}}_{t+1}$, the core task of Sentinel in the pre-scheduling stage is to generate a strategy for potential requests and ensure that the next cycle of request scheduling is revenue-optimized.

We represent the Pre-scheduling Strategy with $PS=\{x_{m,i}^e|e\in[1,E],m\in[1,M],i\in[1,I]\}$, where $x_{m,i}^e$ denotes the number of request of category $i$ from location $m$ served by server $e$.  
The pre-scheduling problem can be defined as:
\begin{align}\label{eq:object}
\max \sum_{e=1}^E F_r \left(\sum_{m=1}^M \sum_{i=1}^I \left(\frac{x_{m,i}^e * \mathbf{A}_{e,m,i}* \mathbf{S}_{e,m,i}}{B_e}\right)\right)* \mathbf{D}_e.
\end{align}
\vspace{-1.0em}
\begin{subequations}
\begin{align}\label{eq:st}
& \text { s.t. } x_{m,i}^e \geq 0\ (e\in[1,E], m\in[1,M], i\in[1,I]), \\  
& \sum_{e=1}^E x_{m,i}^e = \hat{r}_m^i\ (m\in[1,M], i\in[1,I]) \label{eq:st2}, \\  
&  \sum_{m=1}^M \sum_{i=1}^N \frac{x_{m,i}^e * \mathbf{A}_{e,m,i}}{B_e} \leq {U_{opt}} \ (e \in [1, E]),\label{eq:mim_limit} 
\end{align}
\end{subequations}where $B_e$ is the bandwidth capacity of server $e$ and Eq. \ref{eq:object} is our objective function to optimize resource utilization across the CCP, 
Eq. 11(a) imposes the constraint on the range of $x$,  Eq. 11(b) ensures that all requests are scheduled, and Eq. 11(c) means the resource by all requests is less than or equal to the available resource.


We use historical data on startup latency and error rates, along with bandwidth utilization, to determine an appropriate $U_{opt}$. 
Calculating the CDF for both metrics, we set $U_{opt}$ as the smaller value corresponding to the 80th percentile:
\vspace{-0.1em}
\begin{align} \label{eq: u_opt}
\ U_{opt} = \min \{U_{Lat_{80}}, U_{Err_{80}}\}
\vspace{-0.6em}
\end{align}
where $U_{Lat_{80}}$ and $U_{Err_{80}}$ correspond to the 80th percentile of the startup latency and error rates, respectively.

Nevertheless, solving Eq. \ref{eq:object} directly is NP-Hard due to it constitutes an Integer Nonlinear Programming (INP) problem with a vast solution space ($E*M*N$ is on the million scale).
To circumvent this issue, we propose a linear integer relaxation method coupled with a mask-aided joint expectation approximation, reformulating it into a tractable linear one:


$(i)$ 
To improve the solution efficiency, we ignore the request location ($x_{m,i}^e\rightarrow \overline{x}_i^e$). However, processing binary $\mathbf{S}$ destroys the original semantic information on service anomalies.
Since $\mathbf{A_{e,m,i}}$ and $\mathbf{S_{e,m,i}}$ are independent, we jointly approximate them, treating $\mathbf{S}$ as the prior for binary sampling: $\overline{\mathbf{A}}_{e, i}=\mathbb{E}[\hat{A}_{e,m,i} \mid (e, i)]$, $\hat{A}_{e,m,i}=A_{e,m,i} \cdot S_{e,m,i}$.
By calculating the expectation from the sampling, we obtain a two-dimensional revenue matrix that preserves service state information and reduces the computational cost of solving $PS$.

$(ii)$ Inspired by \cite{xu2018electric}, we further transform the problem into a linear integer programming problem by 
outer-approximation scheme with $\mathcal{F}_a (\cdot)$.
Specially, we employ the secant-based approach and get the objective function is replaced with:
\begin{align}\label{eq:object1}
\max \sum_{e=1}^E\mathcal{F}_a \left(\sum_{i=1}^I \left(\frac{\overline{x}_i^e * \overline{\mathbf{A}}_{e,i}}{B_e}\right)\right)* \mathbf{D}_e.
\end{align}
The constraints can be similarly simplified.
Finally, we relax the integer constraint on $x$ and use the Branch and Cut (BC) algorithm \cite{cordeau2006branch} to solve the Linear Programming problem.

After obtaining the intermediate result $\overline{PS}$, it is rounded to the nearest integer.
We expand location dimension based on the geographic distribution of $\hat{\mathcal{R}}_{t+1}$ and service status, dividing $\overline{P S} \in \mathbb{R}^{E \times N}$ to $P S \in \mathbb{R}^{E \times M \times N}$ according to the proportion of request in different locations as:
\vspace{-0.1em}
\begin{equation}
x_{m, i}^e= \begin{cases} \frac{\overline{x}_i^e \cdot \sum_{i=1}^I \hat{r}_m^i}{\operatorname{SUM}\left(\hat{\mathcal{R}}_{t+1}\right)}, & \text { if } S_{e, m, i}=1. \\ 0, & \text { if } S_{e, m, i}=0.\end{cases}
\vspace{-0.1em}
\end{equation}
Approximation is essential, given the requirement to ensure the completion of pre-scheduling before actual requests arrive. 


\subsection{Post-scheduling Stage}\label{rescheduling}
Once the pre-scheduling strategy is constructed, Sentinel classifies the actual arriving requests into predefined categories by the clustering model, and matches them with the strategy as $f(\mathcal{R}_{t+1}, PS) \rightarrow S^*$, where $f$ is the matching function, $PS$ is the updated pre-scheduling strategy in strategy pool, and $\mathcal{S}^*=\{s_{m,i}^e|e\in[1,E],m\in[1,M],i\in[1,I]\}$ is the adapted scheduling strategy. 
As actual deviate from the prediction, 
If actual requests exceed the predictions, the unmatched requests will be processed as follows:

Initially, Sentinel processes the remaining unserved requests using CCP's original heuristic scheduling method, which allocates requests to the nearest servers with the highest remaining bandwidth. The remaining bandwidth, $B_e^{\text{remain}}$, is then recalculated based on the adapted scheduling strategy $\mathcal{S}^*$ and the revenue matrix $A$, formulated as $B_e^{\text{remain}} = B_e - \sum_{m=1}^M \sum_{i=1}^N s_{m,i}^e \cdot A_{e,m,i}$. 
Upon completing this heuristic scheduling, Sentinel effectively schedules all actual requests, thereby completing the entire scheduling cycle.

During long-term operation, it is notable that Sentinel continuously updates its components, including the clustering model, the service effect prediction model, the device anomaly detection model and revenue metric with new $U_{opt}$.

\begin{figure*}[t]
\centering
\includegraphics[width=\textwidth]{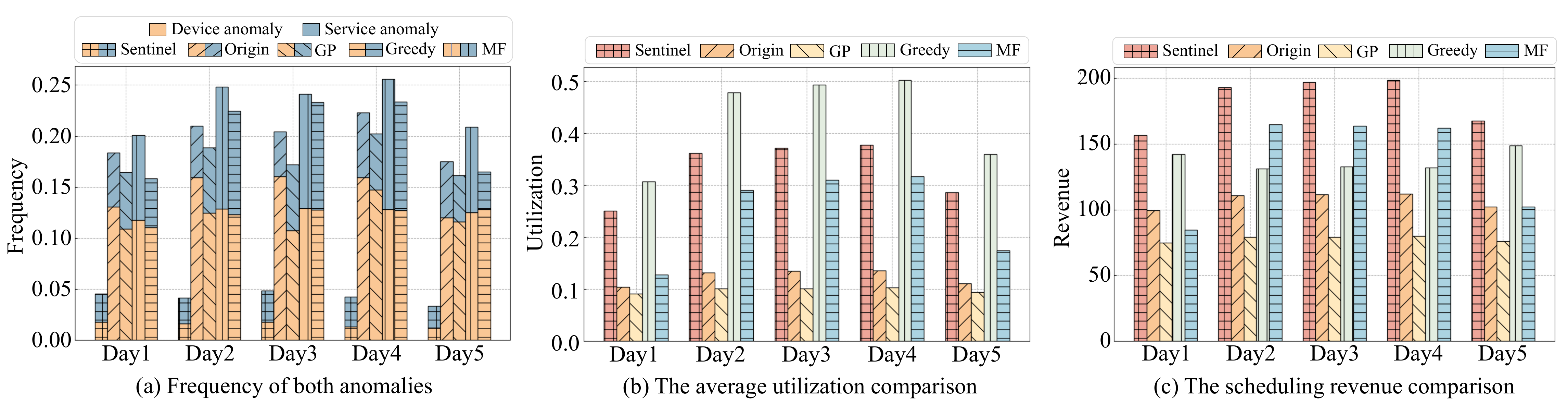} 
\vspace{-1.8em} 
\caption{Scheduling performance comparison\protect\footnotemark[1].}
\label{fig:SchedulingPerformance}
\vspace{-1.2em} 
\end{figure*}
\footnotetext[1]{As the time comes to day2 (weekend), the scale of requests increases significantly.}



\section{PERFORMANCE EVALUATION}\label{EVALUATION}

\subsection{Experiment Setup} \label{exp_setting}
\subsubsection{Datasets and Implementation Details}
With the assistance of a leading CCP in China, our evaluation is based on three distinct real-world datasets, as detailed in Table \ref{table:log_data}.

\begin{table}[ht]
\centering
\begin{threeparttable}
\caption{Summary of Log Data}
\label{table:log_data}
\begin{tabular}{@{}l|p{5.5cm}@{}}
\toprule
\textbf{Log Type} & \textbf{Description} \\
\midrule
\multirow{5}{*}{\makecell[l]{Live Streaming\\Service Logs\\(5.09 GB)}}  
                    & This dataset includes 17,854 live streaming channels and 476 unique server IDs, covering 500 million requests across 59 locations. Data size and error rates are recorded every minute, tracking attributes such as workload, location, and device anomaly. \\
\midrule
\multirow{5}{*}{\makecell[l]{Server-side Logs\\(3.53 GB)}} 
                    & Every 60 seconds, the server records both the successfully served requests and the failed requests due to server overloads or timeouts. The focus is on the error rates of failed requests to reflect service anomalies. \\
\midrule
\multirow{3}{*}{\makecell[l]{Client-reported Logs\\(1.24 GB)}} 
                    & Clients report the total number of requests issued every 60 seconds, including metrics like startup latency, which reflects service anomalies. \\
\bottomrule
\end{tabular}
\end{threeparttable}
\end{table}
We analyzed data from the first five days for CCP measurements and Sentinel setup verification, while using the last five days to evaluate scheduling performance. During this period, the platform experienced an average anomaly frequency of 16.1\%, with device anomalies constituting 30.6\% and service anomalies 55.2\% of all anomalies.

\textbf{System settings.}
We set 10 device clusters and each cluster
contains 5 virtual machines (VMs) acting as 5 devices. 
The test clusters are built from the CCP to match the real heterogeneous situation.
We select devices from the clusters to serve as decision server and implement updates to the policy pool by maintaining a stack. 
To ensure the fairness, Sentinel is executed on the same decision server as all baselines, with scheduling effects collected from the same test cluster.

\textbf{Parameters settings.}
We monitor four key workloads—upload bandwidth, memory, CPU, and disk—for device state.
We set $T$ as $12$, $\beta$ as 50\% of all dimensions, and both $\theta_v$ and $\theta_r$ are set to 0.4.
The initial temperature $\lambda$ in Gumbel-Softmax is 5.0, and anneal to 0.1. The device anomaly threshold $\eta = 0.3$.
For the service effect prediction, we set the request cluster number $k=29$.


\subsubsection{Baselines}
\begin{itemize}[leftmargin=*]
    \item Heuristic method (Origin): This method schedules requests to the nearest servers with maximum remaining bandwidth. Its the origin scheduling method of the collaborated CCP, and we do not perform additional replication but only record the relevant metrics under the real CCP.
    \item Geographically-Proximate (GP): This approach schedules requests to the nearest available  edge servers. It is he most prevalent method in industry. 
    \item Revenue-aware Greedy (Greedy) \cite{Haouari2019}: This method is based on the request revenue matrix $\mathbf{A}$. It allocates requests to the server that yields the highest revenue until the server's bandwidth capacity is reached. It can be regarded as the upper-bound in the context of utilization.
    \item Maximum-flow (MF) \cite{9488868, 8241883}:. This kind of algorithms convert the scheduling problem into a flow control problem (the server is treated as the graph node, the constrained bandwidth is treated as the link capacity, and the scheduled request revenue is treated as the flow).
\end{itemize}

\subsection{Scheduling Performance}  
To evaluate the system-level performance of Sentinel and the baseline methods, we compare them over the last five days.

\subsubsection{Sentinel Effectively Circumvents Anomaly}
Fig. \ref{fig:SchedulingPerformance}(a) illustrates the anomaly frequency for different scheduling methods.
Sentinel demonstrates significant advantages, the average total frequency of two anomalies is less than 0.05, even in such an unstable CCP.
Sentinel reduces ineffective scheduling by an average of 70\% compared to other baselines.
Sentinel directs requests to reliable servers by filtering out those likely to cause scheduling issues based on anomaly detection.

\subsubsection{Sentinel Yields More Balanced Utilization} 
Fig. \ref{fig:SchedulingPerformance}(b) shows that Sentinel achieves the highest reasonable utilization (second only to Greedy's upper-bound),  
while the Greedy method exhibits a higher utilization owing to its inherent scheduling mechanism that favors servers with the highest potential revenue. 
However, ignoring anomaly disturbances to revenue efficiency often leads to the Greedy method surpassing the optimal utilization $U_{opt}$, thereby diminishing its revenue.

When scheduling requests, Sentinel leverages $\mathcal{F}_r$ to dynamically evaluate the potential revenue for each server, and strategically schedules the requests to servers which the utilization remains within the optimal range.

\subsubsection{Sentinel Exhibits Superior Revenue}  
As shown in Fig. \ref{fig:SchedulingPerformance}(c), Sentinel consistently outperforms all baselines across the testing days. Specially,
Sentinel achieves an enhancement of 70\%, 134\%, 17\%, and 76\% in the average revenue comparing to original heuristic method, GP, Greedy and MF, respectively.



These results imply the following conclusions:
$(i)$ All baselines trigger anomaly frequently, which demonstrates the importance of anomaly detection in scheduling.
Notably, the Sentinel further integrates anomaly detection and revenue-aware considerations into present decisions,
effectively preempting  potential  ineffective scheduling situations, which ultimately leads to its superior performance.
$(ii)$ Despite the superiority of Sentinel, it is not entirely immune to anomaly due to the detection errors and the approximation in \(\mathrm{P^2}S\) paradigm. 
However, these occurrences are minimal, reflecting Sentinel's optimal balance between effective anomaly detection and maintaining system efficiency.


\subsection{Time Efficiency}  

\begin{figure}
    \centering
    \includegraphics[width=7.2cm]{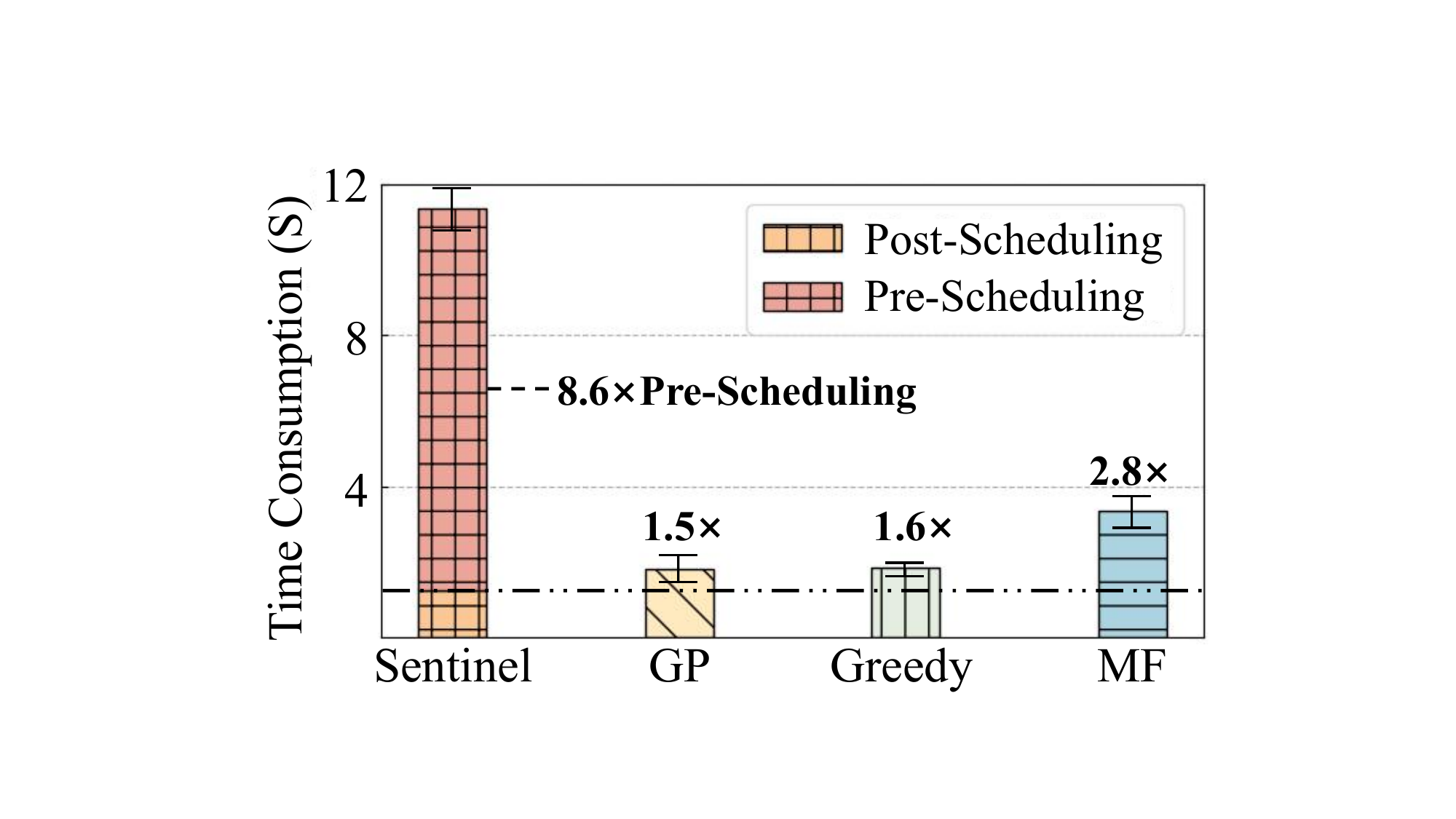}
    \vspace{-1.0em} 
    \caption{Average scheduling time consumption.}
    \label{fig:time_compare}
    \vspace{-1.5em} 
\end{figure}

We evaluate the average time consumption per scheduling cycle for different methods.
As shown in Fig. \ref{fig:time_compare}, Sentinel demonstrates optimal efficiency, highlighting the practicality of the proposed  \(\mathrm{P^2S}\) paradigm for real-time-sensitive scheduling.
Sentinel strategically shifts the time-consuming anomaly detection and scheduling optimization steps prior to the arrival of real-time requests, 
leaving the efficient post-scheduling stages for the actual scheduling cycle.
Sentinel reduces the scheduling time consumption by $1.5\times$, $1.6\times$, and $2.8\times$ compared to GP, Greedy, and MF, respectively.

Current state-of-the-art scheduling methods (e.g., \cite{zhang2019livesmart, 9047133}) fail to fully exploit predictive capabilities, often tailoring the scheduling cycle to optimize specific goals at the cost of significantly increased user-perceived latency. For instance, if Sentinel were to perform anomaly detection and scheduling simultaneously during real-time request arrivals, the total scheduling time would increase by a factor of $8.6 \times$, even with our implementation of efficient detection methods. 

\subsection{Robustness Analysis with Different Anomaly Frequencies}
The proliferation of anomalies in realistic scenarios can significantly impact scheduling, thus we explore the robustness by scheduling for distributions of heterogeneous anomalous devices and time periods with varying service anomaly ratios.

\begin{figure}[t]
    \centering
    \includegraphics[width=\columnwidth]{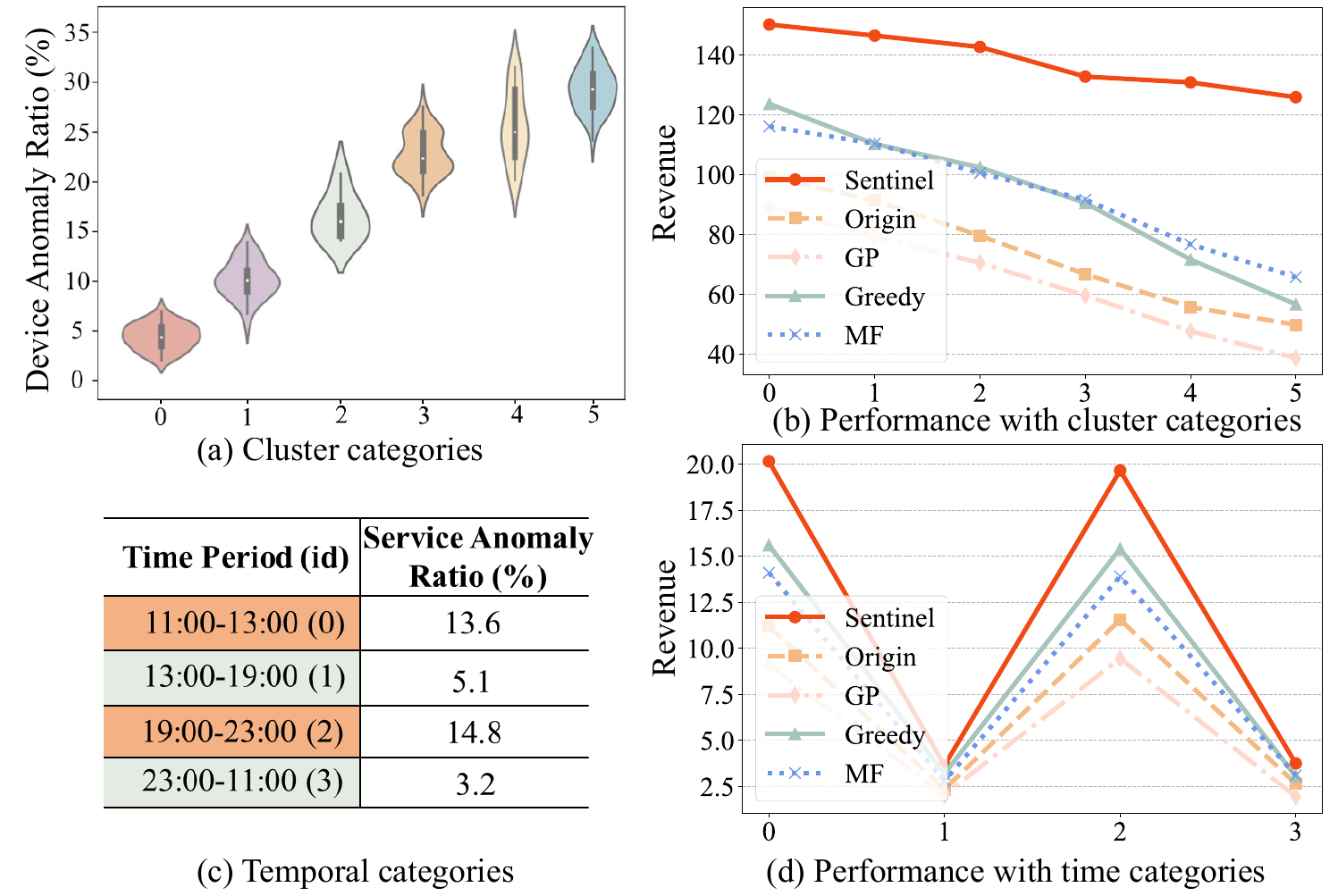}
    \vspace{-2.0em}  
    \caption{Scheduling performance under various anomaly conditions.}
    \vspace{-1.5em}  
    \label{fig:ro}
\end{figure}

Specifically, we categorize all servers on the platform based on their historical failure frequencies. 
As shown in Fig. \ref{fig:ro}(a), servers with smaller cluster id  are more stable and exhibit lower failure rates. 
Fig. \ref{fig:ro}(b) exhibits the scheduling performance for different
clusters. Sentinel surpasses other baselines by a significant margin, which becomes more pronounced as the number of anomalous devices increases.
Next, we partition one day into four time periods, whose categories are given in Fig. \ref{fig:ro}(c). 
 Fig. \ref{fig:ro}(d) resents the hourly scheduling performance. 
Sentinel beats the baselines in terms of every category. 
Furthermore, Sentinel shows significant improvement in categories 0 and 2, which denote peak hours. During these times, the scale of requests increases significantly, followed by a rise in the service anomaly ratio.


Sentinel effectively handles these situations by integrating advanced anomaly detection into real-time scheduling for each cycle, adapting to varying levels of request and anomaly ratios to ensure optimal performance.
This indicates that when CCPs suffers from excessive pressure or sudden failures, Sentinel can still maximize the stability of the service to ensure the revenue, demonstrating strong robustness.

\subsection{Impact of Key Components}  
\begin{figure}[t]
    \centering
    \includegraphics[width=9cm]{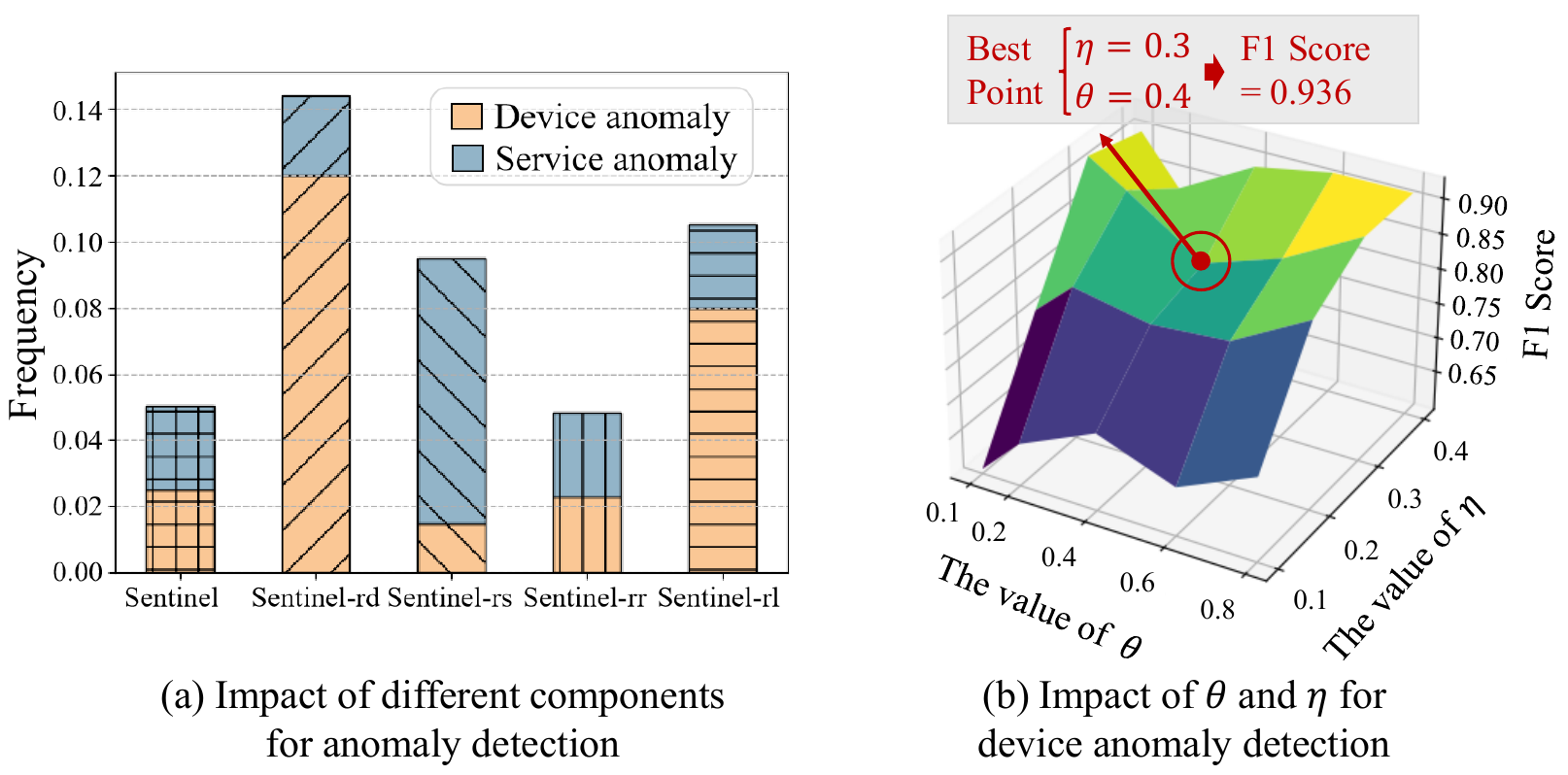}
    \vspace{-1.5em} 
    \caption{Variation in anomaly detection performance with different settings.}
    \label{fig:ablation study}
    \vspace{-1.5em} 
\end{figure}

To get a deeper understanding of Sentinel's key components, we perform ablation studies with five variants as shown in Fig. \ref{fig:ablation study}(a).
We observe that Sentinel significantly reduces the frequency of specific anomalies compared to Sentinel-rd (Removal of Device anomaly detection) and Sentinel-rs (Removal of Service anomaly detection), indicating the effectiveness of our integrated anomaly detection in enhancing the stability and reliability of CCPs.

Moreover, Sentinel outperforms Sentinel-rl (Removal of Learning-based device anomaly detection), which justifies the effectiveness of learning-based probabilistic
mixture detection model.
Although Sentinel-rr (Removal of Rule-based device anomaly detection) performs similarly to Sentinel, its significant time consumption (the total scheduling time would inflate by $2.1\times$) prevents it from completing pre-scheduling before the actual requests arrive. This results from the lack of fast rule-based filtering, which leaves more computational dimensions for subsequent learning-based detection
The findings suggest the necessity of two-stage device anomaly detection for large-scale CCPs, that makes a good trade-off between accuracy and efficiency.

Fig. \ref{fig:ablation study}(b) depicts the influence of varying $\theta$ (both $\theta_v$ and $\theta_r$) and $\eta$ thresholds on device anomaly detection, evaluated by F1 Score \cite{elmrabit2020evaluation}. When $\theta$ and $\eta$ are set to lower values, the detector becomes more sensitive to anomalies. This high sensitivity results in more false positives, reducing the F1 Score. Conversely, when the thresholds $\theta$ and $\eta$ are set to higher values, the system's sensitivity to anomalies decreases, leading to a lower F1 Score as some true anomalies are missed. Notably, when $\theta$ exceeds 0.4, changes in the F1 Score become negligible, indicating that the first stage has been fully utilized. Under this condition, finding a suitable $\eta$ helps balance accuracy and efficiency while minimizing false alarms.

\section{RELATED WORK}\label{rw}

Optimization strategies for live streaming service platforms have garnered extensive attention from both academia and industry in recent years. Compared to modifying architectures and protocols, these optimization strategies have achieved significant improvements at lower costs \cite{barakabitze2019qoe}. Current research primarily focuses on improving QoS \cite{song20244d}, reducing latency \cite{sun2021towards}, and efficiently managing bandwidth \cite{cui2020tclivi}.

Traditional scheduling methods mainly rely on edge caching and location-aware scheduling, which neglect request characteristics and factors like QoS and QoE \cite{turkkan2022greenabr}. To address these issues, some studies have explored deep learning-based methods for optimizing scheduling \cite{zhang2020leveraging,ji2023adaptive}.
Additionally, numerous methods based on deep reinforcement learning frameworks have been proposed for real-time request decision-making \cite{huang2018qarc, naresh2023ppo, tian2019deeplive}.


Although existing methods are effective in certain aspects, they fail to comprehensively consider factors impacting revenue and overlook the importance of anomaly detection \cite{he2023online}. 
Various anomalies, such as unexpected server failures, can occur in streaming services, diminishing the streaming experience and consequently affecting QoS. 
Current anomaly detection methods based on edge cloud platforms, when used to guide scheduling, may introduce significant delays \cite{ngo2020adaptive, li2023digital}, handling of these anomalies particularly challenging in time-sensitive scheduling processes.

\section{CONCLUSION}
This paper introduces Sentinel, a proactive anomaly detection-based scheduling framework for live streaming services in CCPs. Inspired by our meticulous measurements of real-world CCP environments, by integrating a novel \(\mathrm{P^2S}\) paradigm and 
meticulously designed anomaly detection, Sentinel exhibits proactive revenue-optimized scheduling in large-scale CCPs. Over five days of testing, Sentinel demonstrated superior performance, boosting CCP origin revenue by 74\% and accelerating scheduling $2.0\times$ faster than its counterparts.


\section*{Acknowledgement}
This work was supported in part by the Tianjin Natural Science Foundation General Project under Grant No. 23JCYBJC00780, and the National Science Foundation of China under Grant No. 62072332.
Thanks to PPIO Cloud Computing Co., Ltd for supporting this work. 








\bibliographystyle{IEEEtran}


\end{document}